%% file: main_arxiv.tex
\documentclass[final]{cvpr}

\usepackage{times}
\usepackage{epsfig}
\usepackage{graphicx}
\usepackage[dvipsnames]{xcolor}
\usepackage{amsmath}
\usepackage{amssymb}
\usepackage{soul}
\usepackage{booktabs}
\usepackage{dirtytalk}
\usepackage[numbers,sort]{natbib} 
\usepackage[dvipsnames]{xcolor}
\usepackage[per-mode=fraction]{siunitx}
\usepackage{balance}

\def\sepappendix{0}

\newcommand{\bslcorpus}{\textsc{BslCorpus}}

\newcommand{\bslonek}{BSL-1K}
\newcommand{\methodName}{CMPL}
\newcommand{\methodNameLong}{Changepoint-Modulated Pseudo-Labelling}

\newcommand{\pseudolabel}{pseudo-label}
\newcommand{\pseudolabels}{pseudo-labels}
\newcommand{\pseudolabelling}{pseudo-labelling}
\newcommand{\Pseudolabel}{Pseudo-label}
\newcommand{\Pseudolabels}{Pseudo-labels}
\newcommand{\Pseudolabelling}{Pseudo-labelling}

\newcommand{\latestEdit}[1]{{#1}}

\usepackage[pagebackref=true,breaklinks=true,colorlinks,bookmarks=false]{hyperref}

\def\cvprPaperID{2} 
\def\confYear{SLRITW 2021}

\begin{document}

\title{Sign Segmentation with \methodNameLong}

\author{
Katrin Renz\textsuperscript{1,2}
\quad
Nicolaj C. Stache\textsuperscript{2}  
\quad
Neil Fox\textsuperscript{3}
\quad
 G\"ul Varol\textsuperscript{1,4}
\quad
Samuel Albanie\textsuperscript{1}
 \\
{\normalsize \textsuperscript{1}Visual Geometry Group, University of Oxford, UK}\\
{\normalsize \textsuperscript{2} University of Heilbronn, Germany}\\
{\normalsize \textsuperscript{3} Deafness, Cognition and Language Research Centre, University College London, UK}\\
{\normalsize \textsuperscript{4} LIGM, \'Ecole des Ponts, Univ Gustave Eiffel, CNRS, France}\\
{\tt\small \url{https://www.robots.ox.ac.uk/~vgg/research/signsegmentation}}
}


\maketitle

\input{abstract.tex}
\input{intro.tex}
\input{related.tex}

\input{method.tex}
\input{experiments.tex}
\input{conclusions.tex}

\input{acknowledgements.tex}

{\small
\bibliographystyle{ieee_fullname}
\bibliography{refs}
}

\clearpage

\bigskip
{\noindent \large \bf {APPENDIX}}\\
\input{appendix.tex}

\balance

\end{document}

%% file: abstract.tex
\begin{abstract}
The objective of this work is to find temporal boundaries between signs in continuous sign language.
Motivated by the paucity of annotation available for this task, we propose a simple yet effective algorithm to improve segmentation performance on unlabelled signing footage from a domain of interest.
We make the following contributions:
(1) We motivate and introduce the task of source-free domain adaptation for sign language segmentation, in which labelled source data is available for an initial training phase, but is not available during adaptation.
(2) We propose the \methodNameLong{} (\methodName) algorithm to leverage cues from abrupt
changes in motion-sensitive feature space to improve \pseudolabelling{} quality for adaptation.
(3) We showcase the effectiveness of our approach for category-agnostic sign segmentation, transferring from the \bslcorpus{} to the \bslonek{} and RWTH-PHOENIX-Weather 2014 datasets, where we outperform the prior state of the art.
\end{abstract}

%% file: intro.tex
\section{Introduction}
\label{sec:intro}

Sign languages are visuo-gestural, evolved languages that represent the natural means of communication for deaf communities~\cite{sutton-spence_woll_1999}. Automatic systems for recognising and understanding signing content have a wide range of applications: enabling indexing of signing content to facilitate efficient search, assistive tools for education and sign linguistics analysis, and sign \say{wake-word} recognition for virtual assistants~\cite{rodolitz2019accessibility,bragg2019sign}.

A major challenge in developing such systems is the relative paucity of annotated sign language data that may be employed for training~\cite{bragg2019sign,koller2020quantitative}. Several factors drive this state of affairs: a limited supply of annotators with requisite knowledge of sign language required to perform labelling and the extremely high cost of producing the labels themselves~\cite{Dreuw2008TowardsAS}.

\begin{figure}
    \centering
    \includegraphics[width=0.47\textwidth,clip,trim={23cm 20cm 27cm 3cm}]{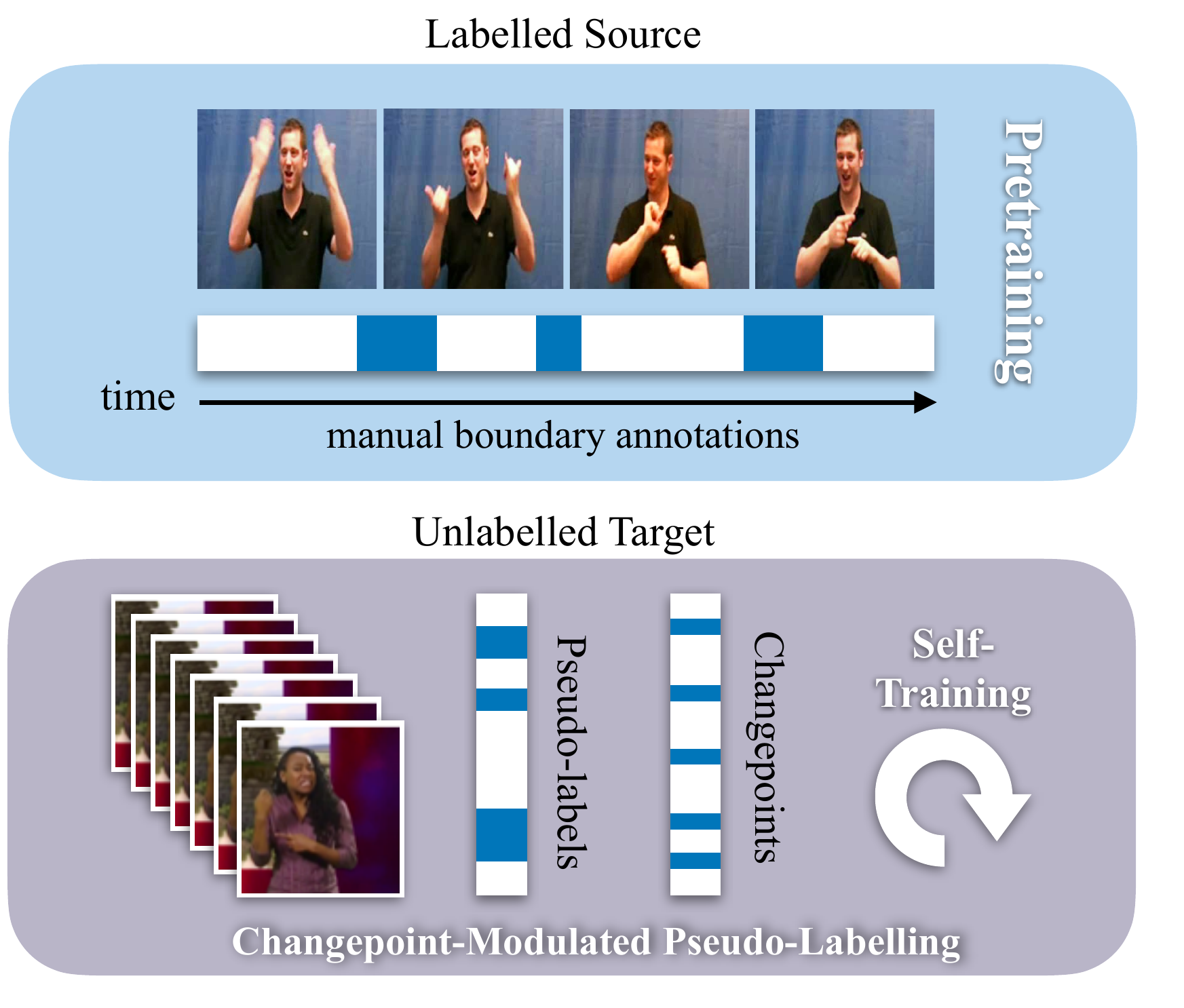}
    \caption{\textbf{\methodNameLong{} for sign segmentation:} Following an initial pretraining phase on annotated data (top), our proposed approach leverages changepoints detected in motion-sensitive feature space to enhance the quality of annotation produced by \pseudolabelling{} video data on the target domain of interest (bottom) to perform \textit{source-free adaptation} (target domain adaptation without concurrent access to the source domain).
    Sign boundaries are marked with blue bars.}
    \mbox{}\vspace{-0.4cm} \\
    \label{fig:teaser}
\end{figure}

In this work, we focus our attention on the task of sign segmentation and propose an automatic method for temporally localising sign boundaries in videos of continuous sign language. Sign segmentation plays an important role in the construction of sign language corpora~\cite{cormier2016digging,cormierbsl} and therefore the development of tools that can perform this task automatically offer the potential to alleviate the limited supply of labelled corpora currently available. 

A number of factors make the objective of localising boundaries in continuous sign language challenging. To solve the task, a model must be capable of discriminating gestural patterns from hands moving at high speed, typically in the presence of motion blur, as well as subtle cues in other modalities such as facial expressions. 
Moreover, in contrast to problem \latestEdit{settings} such as human action segmentation which have profited from a diverse collection of densely annotated datasets~\cite{fathi2011learning,stein2013combining,kuehne2014language,yeung2015every,damen2018scaling,gao2017tall}, existing sign language datasets provide a relatively small quantity of data with precise temporal boundaries~\cite{bslcorpus17,jahn2018publishing}, limiting their ability to train models that can generalise robustly beyond the training domain.

An additional factor of particular relevance to sign language datasets relates to the need to protect the privacy of individuals who have contributed human data. In particular, while marginalized communities may potentially benefit from systems (such as those for automatic sign language understanding) that are designed to meet their needs, they face higher risk by contributing data to those systems~\cite{bragg2020exploring}.   This is because smaller group size makes personal identification easier, and marginalized status renders any privacy breaches that occur more dangerous.

In this work, we therefore consider the setting in which the signing data to be segmented, the \say{target} domain, is sourced from a different distribution to the videos for which segmentation annotation was available for training the model, the \say{source} domain \latestEdit{(see Fig.~\ref{fig:data} for examples of different domains considered in this work)}. We further assume that while the source domain data is available for training an initial model, it is no longer accessible when adapting to the target domain, reflecting the assumption that it may be feasible to share models trained on human data, but infeasible to share the original training data itself.

To make use of the unlabelled data that is available on the target domain, we propose an approach inspired by the classical technique of \textit{\pseudolabelling{}}, in which a classifier is retrained on its own predictions on unlabelled data to improve performance~\cite{lee2013pseudo}. This method, which has proven popular in the context of semi-supervised learning, leverages implicitly the \textit{cluster assumption}---namely that the decision boundaries of the model should lie in regions of low density~\cite{chapelle2009semi} to adapt the model to new examples. While this assumption is reasonable in the absence of other knowledge, it can be problematic when transferring a sign segmentation model from one domain to another when invariances learned on the former are inappropriate on the latter. Particularly for a small source training set, there may be forms of sign boundaries to which the segmentation model has simply not been exposed, and the discovery of such examples through entropy minimisation schemes~\cite{grandvalet2006entropy} such as \pseudolabelling{} is unlikely.

Consistent with this hypothesis, we show through experiments in Sec.~\ref{sec:experiments} that \pseudolabelling{} provides a boost to performance but exhibits a subtle systematic bias towards under-segmentation. To address this issue, one finding of this work is that it is possible to correct this bias by encoding a simple assumption into the \pseudolabelling{} process, namely that \textit{sign boundaries typically correspond to motion disfluencies}. To this end, we propose an extension to \pseudolabelling{} that uses changepoint detection among sequences of motion-sensitive features to modulate the labelset produced on the target domain, increasing their sensitivity to abrupt changes in feature space \latestEdit{(see Fig.~\ref{fig:teaser})}.

We make three contributions:
(1) We motivate and introduce the task \textit{source-free} domain adaptation for sign language segmentation, in which labelled source data is available for an initial training phase, but is not available during adaptation. To the best of our knowledge, despite its importance, this task has not been previously investigated in the literature.
(2) We establish baselines for this task and propose a modification to \pseudolabelling{}, which we term \textit{\methodNameLong{}}, to address the under-segmentation bias exhibited by naive \pseudolabelling{} on the target domain.
(3) We showcase the effectiveness of our approach for category-agnostic sign segmentation, transferring from the \bslcorpus{}~\cite{bslcorpus17,schembri2013building} to the \bslonek{}~\cite{albanie20_bsl1k} and RWTH-PHOENIX-Weather 2014~\cite{Koller15cslr} datasets, where we outperform the prior state of the art. 

%% file: related.tex
\section{Related Work}
\label{sec:related}

Our work relates to several themes that have been explored in the literature: \textit{temporal action segmentation}, \textit{\pseudolabelling{} techniques}, \textit{source-free domain adaptation}, 
\textit{changepoint detection} and \textit{sign language segmentation}.

\vspace{0.1cm}
\noindent \textbf{Action segmentation.} The temporal segmentation of untrimmed videos into sequences of actions has received a great deal of attention in the literature, leading to the development of temporal sliding window classifiers~\cite{rohrbach2012database,karaman2014fast}, generative techniques with Hidden Markov Models (HMMs)~\cite{kuehne2016end,tang2012learning} and stochastic grammars~\cite{vo2014stochastic}. More recently, the considerable effectiveness of temporal convolutional networks has been demonstrated for action segmentation, notably with the introduction of Multi-Stage Temporal Convolutional Networks~\cite{Farha_2019_CVPR} (which we employ in our approach). 
Of relevance to our approach, the work of~\cite{chen2020action} explored the task of domain adaptation for action segmentation networks. While effective, their approach requires continued access to the source domain to enable feature alignment between domains (in common with many other unsupervised domain adaptation methods~\cite{ben2010theory,ganin2016domain,volpi2018adversarial,pei2018multi}), and is therefore not applicable in our problem formulation.

\vspace{0.1cm}
\noindent \textbf{\Pseudolabelling.} The use of \pseudolabelling{} schemes for exploiting unlabelled data to improve performance has a long history of study stretching back to the 1960s~\cite{scudder1965probability,yarowsky1995unsupervised}. Variants of this idea have been explored in semi-supervised learning, in which the model is assumed to have access to both labelled and unlabelled data from which to learn from~\cite{grandvalet2006entropy,chapelle2009semi}. One particular formulation of this idea known as \textit{\pseudolabelling{}}~\cite{lee2013pseudo} has emerged as an especially effective mechanism for semi-supervised learning in which the predictions of a classification model are discretized into one-hot categorical labels and assigned to unlabelled examples, then mixed in with labelled examples to provide an updated training set for the model.
\latestEdit{A form of \pseudolabelling{} was also recently considered for the task of sign recognition by~\cite{li2020transferring} who proposed to localise additional training samples in unlabelled news footage to achieve greater robustness.}
We employ \pseudolabelling{} in the specialised \textit{source-free} setting in which no labelled examples are available after the initial training phase on the source domain, discussed next.

\vspace{0.1cm}
\noindent \textbf{Source-free domain adaptation.}
The adaptation of a model 
to unlabelled data on a target domain of interest without concurrent 
access to labelled source domain data becomes more important due to privacy policies in certain domains. Recently the task has been explored using generative models~\cite{hou2020source, kundu2020universal, li2020model}, with class prototypes~\cite{yang2020unsupervised} or \pseudolabelling{}~\cite{liang2020we}.
In this work, we also use \pseudolabelling{} to adapt models but in addition, we seek to explicitly encode knowledge about the sign segmentation task through changepoint detection.


\vspace{0.1cm}
\noindent \textbf{Changepoint detection.} 
Our category-agnostic sign boundary detection naturally
relates to changepoint detection algorithms~\cite{10.1093/biomet/42.3-4.523,darkhovski,Chen:1498647}, in which
the goal is to locate state changes in time series
in order to segment the underlying signal.
We refer the reader to~\cite{TRUONG2020107299}
for a detailed overview of offline changepoint detection
methods. In this work, we propose a learning framework
in which we integrate the bottom-up changepoint detections
in our high-dimensional video features
obtained from motion-sensitive pretraining tasks.

\vspace{0.1cm}
\noindent \textbf{Sign segmentation.} The segmentation of continuous streams of signing into individual sign \say{tokens} has been the subject of considerable interest in the sign linguistics community, with a particular focus on how to define appropriate boundaries~\cite{cormierbsl,crasborn2008corpus,hanke2012} and measuring consistency between annotation teams~\cite{braffort2012defi,gonzalez2012computer}.
By contrast, automatic sign segmentation has received relatively limited attention in the computational literature. Of the prior research in this area, the dominant approach has been to employ methods that require a semantic labelling of the signing content 
(this can take the form of glosses---the minimal lexical units of signing used for annotation, or full translations)~\cite{santemiz2009automatic,koller2017re}. \latestEdit{This differs from continuous sign language recognition (CSLR)~\cite{koller2015continuous} which aims to determine sign order without localising boundaries (note that while such methods may produce sign boundaries implicitly, they likewise require access to dense semantic labels)}. Other methods have proposed to tackle related tasks such as identifying whether a person is actively signing~\cite{borg2019sign,moryossef2020real} and parsing continuous sign language into sentence-like units~\cite{bull2020}.
Of most relevance to our work, the recent works of~\cite{farag2019learning} and~\cite{renz20segments} proposed methods to tackle the category-agnostic segmentation problem. In~\cite{farag2019learning}, the authors investigated the use of a random forest with geometric features derived from 3D skeleton data (gathered via motion capture). The later work of~\cite{renz20segments} demonstrated the superiority of the MS-TCN architecture~\cite{Farha_2019_CVPR} for this task, and we therefore adopt this model in our approach. Differently from~\cite{renz20segments} who assume the training and test data are drawn from the same distribution, we consider the setting in which the target domain differs from source domain.


%% file: method.tex
\section{Method}
\label{sec:method}

We propose a simple adaptation method for improving sign segmentation performance on a target domain for which labelled data is unavailable under the assumption that the source domain (containing labelled data) may not be accessed during the adaptation process (we formalise this problem definition in Sec.~\ref{sec:method:problem}).  Our approach employs a standard fully-supervised training phase on the source domain (Sec.~\ref{sec:method:src}), before running a secondary phase of iterative adaptation with the \textit{\methodNameLong{}} algorithm (described in Sec~\ref{sec:method:combined}).

\subsection{Problem formulation} \label{sec:method:problem}


\noindent Let $\mathcal{X}$ denote the set of videos containing a person performing continuous sign language. The goal of temporal sign segmentation is to produce for each $\mathbf{x} \in \mathcal{X}$ with frames $(x_1, \dots, x_N)$ a corresponding set of frame labels  $\mathbf{y} = (y_1, \dots, y_N) \in \{0, 1\}^N$, in which label values of $1$ and $0$ denote boundaries between sign segments and the interior of segments, respectively.
We assume access to a set of \textit{labelled} sign segments that are gathered from a \say{source} domain $\mathcal{X}_S \subset \mathcal{X}$ on which an initial model may be trained to perform the segmentation task.  We further assume that this domain differs from that of the \say{target} videos  $\mathcal{X}_T \subset \mathcal{X}$ of interest, for which annotations are not available.  The objective of this work is to maximise sign segmentation performance on $\mathcal{X}_T$, under the constraint that at no point we have concurrent access to video samples from both the source and target domain.

\subsection{Source domain training} \label{sec:method:src}

To make best use of labelled data in the source domain, we first train a sign segmentation model in a fully-supervised manner.  Following the dominant approaches in the temporal segmentation literature~\cite{Farha_2019_CVPR,li2020ms}, we assume that the sign segmentation model decomposes into a visual feature extractor, $\phi(\mathbf{x})$, (typically instantiated as a spatio-temporal convolutional network such as I3D~\cite{carreira2017quo}) and a segmentation network, $\psi (\phi)$ which ingests these features and outputs per-frame segmentation labels~\cite{Farha_2019_CVPR,renz20segments}. In this work, we employ a standard frame-level cross-entropy loss in combination with the smoothing loss proposed by~\cite{Farha_2019_CVPR} to train $f = \psi \circ \phi$. Once source domain training has completed, the source videos and labels are discarded and only the trained model, $f$, is retained.

\begin{figure}
    \centering
    \includegraphics[width=0.48\textwidth]{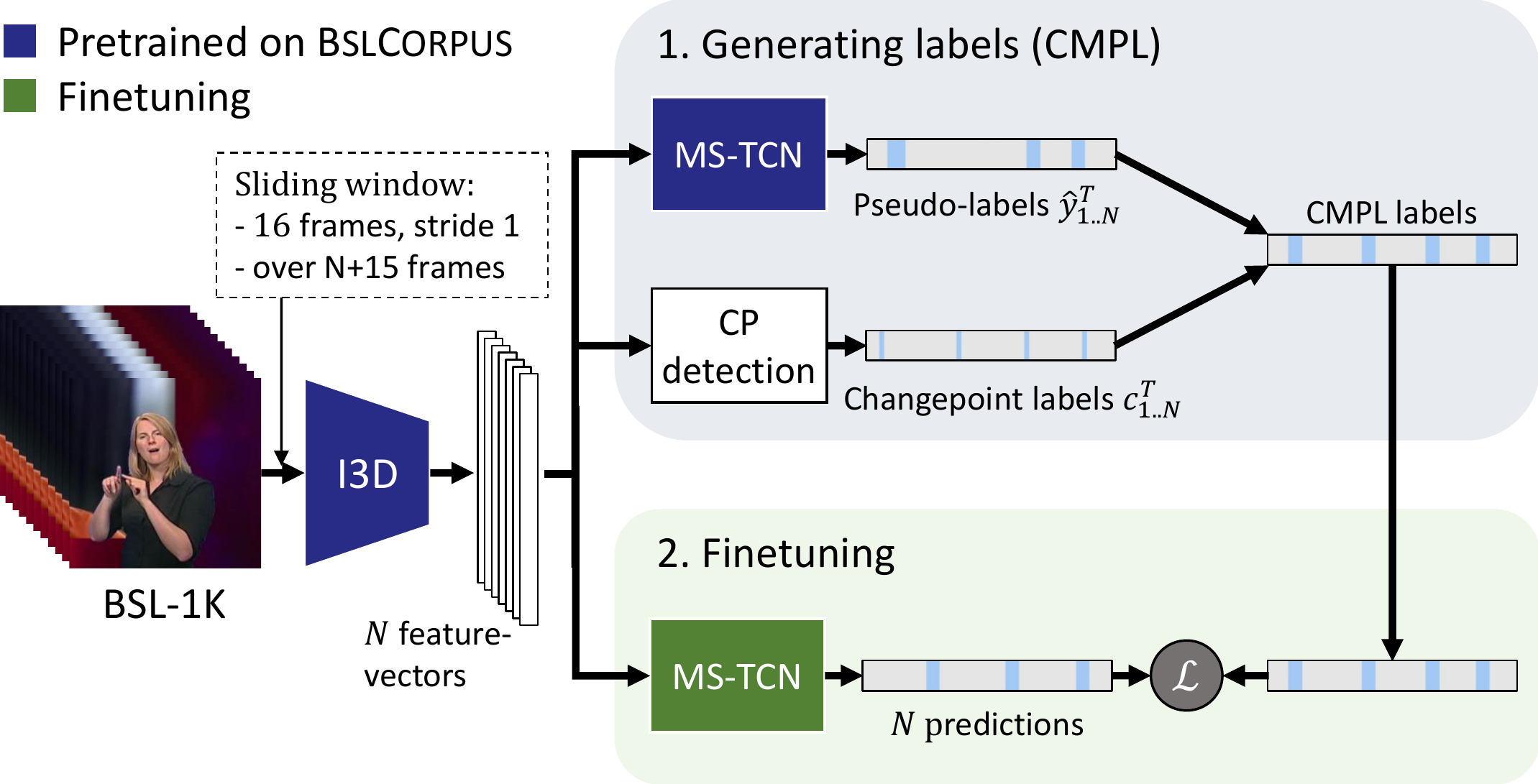}
    \caption{\textbf{Method overview:} The proposed approach extracts I3D features and employs both semantic and low-level cues to produce labels on the target dataset (block 1) via the \methodName{} algorithm.  These labels are then employed directly via finetuning (block 2).
    }
    \mbox{}\vspace{-0.7cm} \\
    \label{fig:blockdiagram}
\end{figure}

\subsection{\methodNameLong{}} \label{sec:method:combined}

The key idea behind our adaptation approach is to leverage the benefits of \textit{\pseudolabelling{}}~\cite{lee2013pseudo} while mitigating its limitations for the task of sign language segmentation through a simple technique we term \textit{changepoint-modulation}. 
A method overview can be seen in Fig.~\ref{fig:blockdiagram}.

\Pseudolabelling{} works by minimising the conditional entropy of a given classification model across unlabelled samples. It achieves this by assigning to each sample the class label corresponding to the maximum posterior probability of the model and then retraining the model to become more confident in its predictions. In doing so, it encodes the intuition that individual samples should not represent a mixture of classes---rather they should belong to one or the other (corresponding to the \textit{cluster assumption} that decision boundaries should lie in regions of low-density~\cite{chapelle2009semi}).
In this work we employ \pseudolabelling{} in the manner described above: we perform inference with the sign segmentation model $f = \psi \circ \phi $ (that was trained on the source domain) across videos of unlabelled signing drawn from $\mathcal{X}_T$ to produce frame-level posterior probabilities for each frame. We then assign each frame the binary label corresponding to the maximum posterior probability of the model and retrain the model by employing the same losses that were used on the source domain (Sec.~\ref{sec:method:src}).

As we show through experiments in Sec.~\ref{sec:experiments} (and consistent with the literature~\cite{lee2013pseudo,zou2018unsupervised,saito2017asymmetric}), \pseudolabelling{} yields a boost in performance on the target domain. However, while it brings consistent improvement, we observe that in practice the algorithm exhibits a subtle but systematic bias towards \textit{under-segmenting} signs.  In the absence of other knowledge, the cluster assumption represents a reasonable prior to improve target domain performance. However, for the particular task of sign segmentation, the sign linguistics literature suggests that there are particular gestural movements and cues that correlate with sign boundaries~\cite{johnson2011segmental,hanke2012} which we can expect to hold true across domains. While prescribing a precise set of rules that define a sign boundary would be extremely challenging~\cite{hanke2012}, we can nevertheless integrate our knowledge about the nature of these boundaries---in particular, their weak correspondence with human motion disfluencies---into the \pseudolabelling{} process. To do so, we propose to couple motion-sensitive features (which may be learned on a host of human activity-based visual understanding tasks) with changepoint detection in feature space \latestEdit{(implementation details are given in detail in Sec.~\ref{subsec:exp:implementation})}. In this way, we obtain, in addition to the \pseudolabels{} described above, an additional set of candidate frame-locations at which a sign boundary is more likely to occur (those that correspond to discontinuities in feature space detected by the changepoint algorithm).

A natural question then arises: \textit{how should the two sets of candidate frame-level labels be integrated to improve target domain performance?} One simple strategy would be to take the union of boundaries predicted by both methods. However, this strategy, as with that of simple averaging, does not account for the fact that intuitively, we would like \pseudolabels{} to be reinforced when they appear in a close neighbourhood of a motion feature changepoint without requiring exact alignment to achieve this effect. 

\begin{figure}
    \centering
    \includegraphics[width=0.47\textwidth]{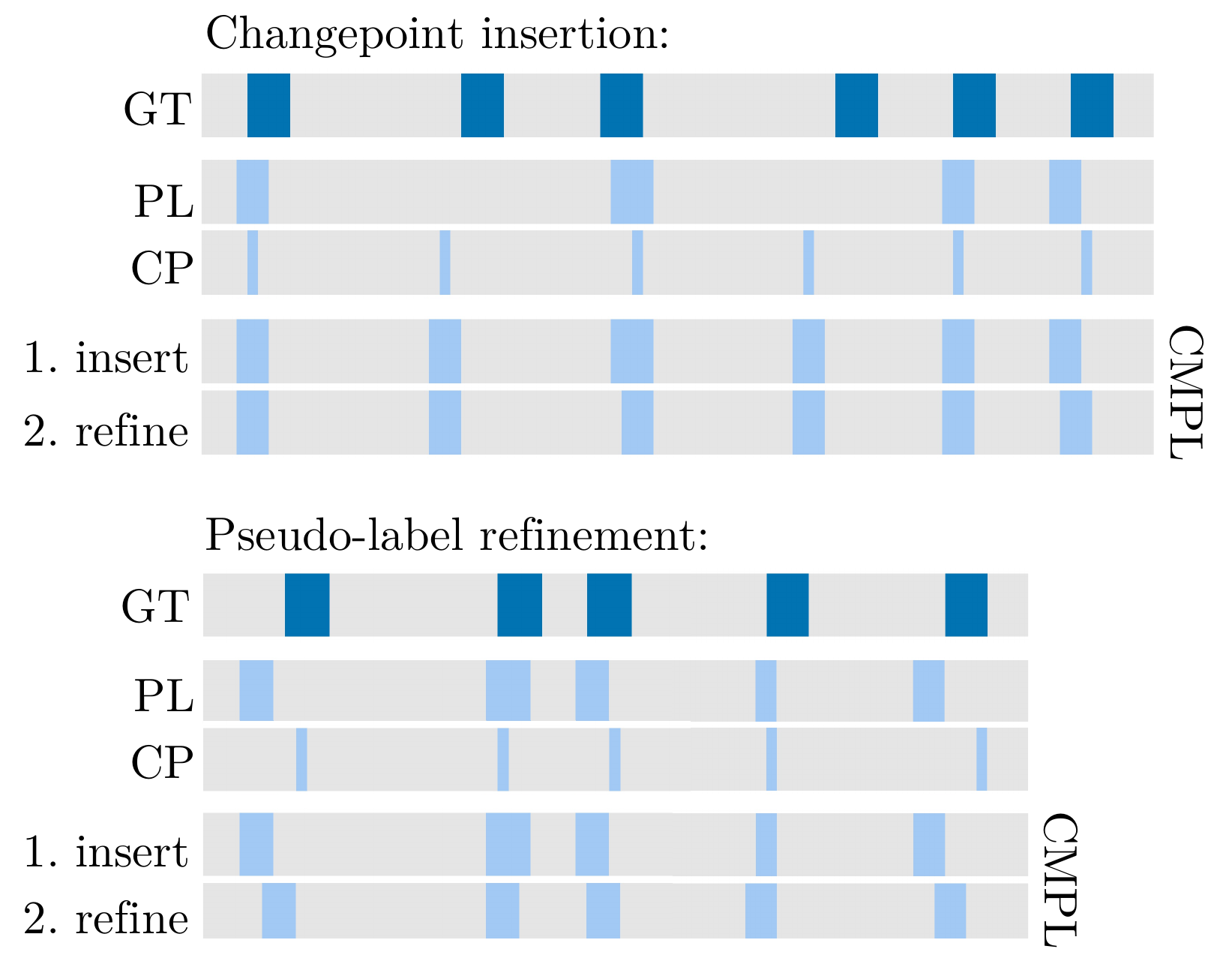}
    \caption{\textbf{\methodName{} \latestEdit{components}:} We show the effect of the insertion and refinement \pseudolabel{} transformations on a sequences of \pseudolabels{} (PL) and changepoints (CP) relative to ground truth (GT) (see Sec.~\ref{sec:method:combined} for details). (Top) An example where the changepoints help to mitigate under-segmentation \latestEdit{in the PL sequence}. (Bottom) Small refinements on the \pseudolabels{} lead to improved boundary locations. 
    }
    \mbox{}\vspace{-0.7cm} \\
    \label{fig:fusion}
\end{figure}

To address this, we propose the \textit{\methodNameLong{}} algorithm, which comprises two \pseudolabel{} transformations. Motivated by the under-segmentation issue described above, the first transformation inserts new boundaries suggested by abrupt changes in feature space, when far from \pseudolabel{} boundaries. More concretely, given a sequence of target-domain frame-level binary \pseudolabels{}, $(\hat{y}_1^T, \dots, \hat{y}_N^T)$, and frame-level binary changepoint labels $(c_1^T, \dots, c_N^T)$ (obtained by placing a unit value at every frame corresponding to a changepoint), we perform the following \pseudolabel{} \textit{insertion} transformation:

\begin{equation}
\hat{y}_{i}^T= 
\begin{cases}
    c_i^T, & \text{if } \sum_{j\in(-\gamma, \gamma)} \hat{y}_{i+j}^T = 0 \\
    \hat{y}_i^T,  & \text{otherwise}
\end{cases}
\end{equation}
where $\gamma$ represents a bandwidth value (set as a hyperparameter). The second \textit{refinement} transformation aims to minimise potential bias towards the annotation style used in the source dataset. First, each contiguous sequence of boundary labels in the \pseudolabels{} is matched to the nearest contiguous sequence of boundary labels in the changepoint sequence, under the condition that it falls within a matching window of $\delta$ frames (also set as a hyperparameter). Matched contiguous \pseudolabel{} boundary sequences are then translated in time such that their central element occupies the midpoint between the original \pseudolabel{} and changepoint boundary positions prior to performing the translation. Examples for these transformation steps can be seen in Fig.~\ref{fig:fusion}.

Following these transformations, the sign segmentation model is simply retrained on the updated set of \pseudolabels{} across the target domain videos.  We conduct ablations to assess the utility of both the \textit{insertion} and the \textit{refinement} transformations in Sec.~\ref{sec:experiments}, as well as the role of the corresponding bandwidth and matching neighbourhood hyperparameters, $\gamma$ and $\delta$.

%% file: experiments.tex
\section{Experiments}
\label{sec:experiments}
This section describes the datasets (Sec.~\ref{subsec:exp:datasets}), the evaluation metrics (Sec.~\ref{subsec:exp:metrics}), and further implementation details (Sec.~\ref{subsec:exp:implementation}). We then present various baselines, ablations and comparisons to the prior state of the art (Sec~\ref{subsec:exp:ablations}). \latestEdit{Finally, we provide
qualitative results (Sec~\ref{subsec:qual}).}

\subsection{Datasets} \label{subsec:exp:datasets}

\label{sec:qualitative}
\begin{figure}
    \centering
    \includegraphics[width=0.48\textwidth]{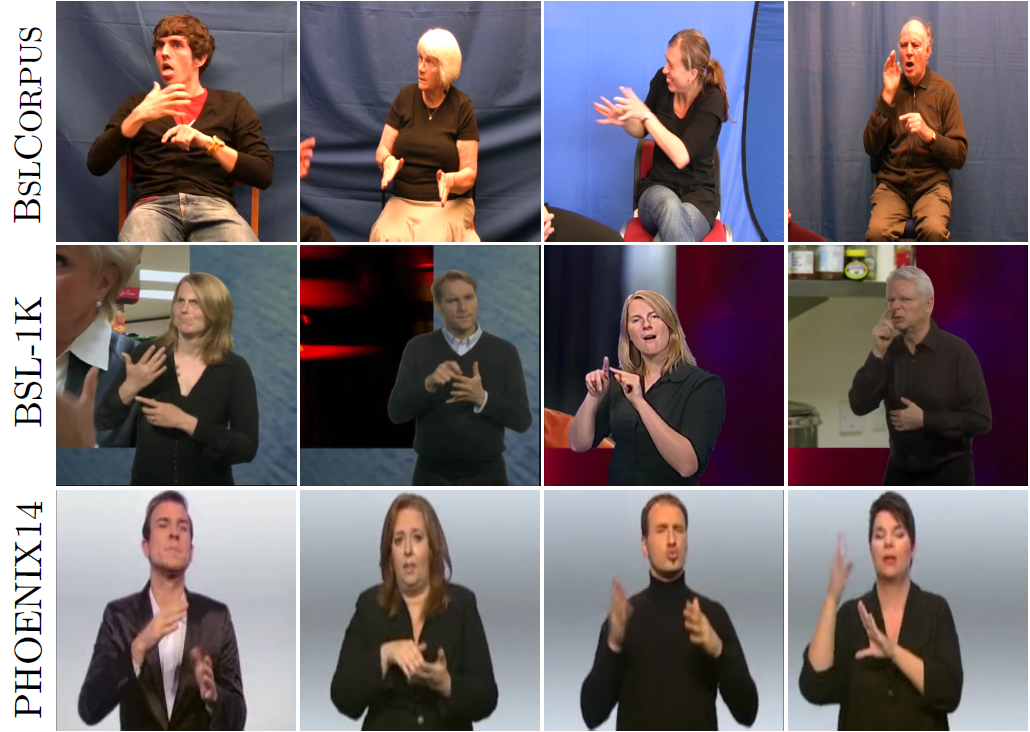}
    \caption{\textbf{Sample frames from the three video datasets considered in this work:} We investigate source-free domain adaptation for sign segmentation across different domains of discourse (personal narratives vs TV broadcast content), context (conversational vs interpreted signing) and sign language (BSL vs DGS). 
    }
    \mbox{}\vspace{-0.7cm} \\
    \label{fig:data}
\end{figure}

We use three datasets in our experiments. Sample video frames from each of these datasets are provided in Fig.~\ref{fig:data}.

\vspace{0.1cm}
\noindent\textbf{\bslcorpus}~\cite{bslcorpus17,schembri2013building} is a British Sign Language (BSL) dataset which contains, inter alia, continuous signing in the form of conversations or narratives. 
For a subset of the data, fine-grained annotations exist which contain start and end times,
as well as the sign categories. 
We use this as the source dataset to 
perform pretraining. We employ the
train and test partitions used in~\cite{renz20segments} 
which contain approximately six hours of signing with gloss-level annotations. A detailed description of the dataset can be found via~\cite{bslcorpus17}.

\vspace{0.1cm}
\noindent\textbf{\bslonek}~\cite{albanie20_bsl1k} consists of
public broadcast footage of sign language interpreted videos, organised into
a number of episodes. 
This dataset shares the same language as \bslcorpus{} (i.e., BSL) but differs in content, background, person position, and signing speed.
To evaluate the performance on this dataset we use the manually-annotated subset of signing data provided by~\cite{renz20segments}, comprising two minutes of video footage with 177 sign instances.
We experiment with training on a subset of three episodes of \bslonek{} corresponding to \SI{3.5}{h} of signing content. We also conduct experiments under a \textit{transductive setting} in which we perform adaptation directly on the episode from which the labelled test set sequence was sourced (this episode represents \SI{45}{min} of signing content). 


\vspace{0.1cm}
\noindent\textbf{RWTH-PHOENIX-Weather 2014}~\cite{Koller15cslr} (PHOENIX14) is a German Sign Language (DGS) dataset that has been widely employed for the study of automatic sign language recognition. The work of \cite{koller2017re} provides automatic temporal segment labels for the training partition of this data, which they obtain through forced-alignment (using ground-truth gloss information). We use these labels against which we evaluate our predictions. To enable direct comparison, we employ the same train/test partitions as used in~\cite{renz20segments} \latestEdit{resulting in a \SI{1.76}{h} test set.}

\subsection{Evaluation metrics} \label{subsec:exp:metrics}
We use the same metrics as in~\cite{renz20segments} and report for each the mean and standard deviation out of three runs with different random seeds.
They use two different metrics: mF1B to measure the performance of the boundary position and mF1S for the extent of the sign segments. \\
\textbf{mF1B.} One boundary, which we define as a series of following $1$s in $y$, is predicted correct, if the distance to a ground truth boundary is lower than a certain threshold. We use the mean of all F1 scores with thresholds in the interval [1, 4]. \\
\textbf{mF1S.} A sign segment is counted as correct if the IoU is higher than a given threshold. We calculate the average F1 score for all thresholds in the range of $0.4$ and $0.75$ with the step size $0.05$. 
This metric is, in contrast to the mF1B metric, sensitive to the width of the predicted and ground truth boundaries, which leads also to sensitivity to different annotation styles between annotators. To avoid this, we consider the mF1B metric as our main metric for ranking the results but include mF1S for completeness.


\begin{table}
    \centering
    \resizebox{0.85\linewidth}{!}{
    \begin{tabular}{lccc}
         \toprule
          & Adaptation protocol & mF1B & mF1S   \\
         \midrule
         Source-only \cite{renz20segments} & & $46.75_{\pm1.2}$ & $32.29_{\pm0.3}$\\
         \midrule
          \Pseudolabels & inductive &  $47.94_{\pm1.0}$ &  $32.45_{\pm0.3}$ \\
          Changepoints & inductive &  $48.51_{\pm0.4}$ &  $34.45_{\pm1.4}$ \\ 
          \methodName{} & inductive & $\mathbf{53.57_{\pm0.7}}$ &  $33.82_{\pm0.0}$\\
         \midrule
         \Pseudolabels{} & transductive & $47.62_{\pm0.4}$ &  $32.11_{\pm0.9}$  \\
         Changepoints & transductive &  $48.29_{\pm0.1}$ &  $35.31_{\pm1.4}$  \\ 
         \methodName{} & transductive  & $\mathbf{53.53_{\pm0.1}}$ &  $32.93_{\pm0.9}$\\
         \bottomrule
    \end{tabular}
    }
    \vspace{6pt}
    \caption{\textbf{Results on the PHOENIX14 dataset:} (Top) Naive sign segmentation performance without assuming any target domain labels. (Middle and Bottom) A comparison of adaptation strategies in the inductive (middle) and transductive (bottom) setting. In each case, we see that the proposed \methodNameLong{} method (\methodName) outperforms alternatives.
    }
    \mbox{}\vspace{-0.7cm} \\
    \label{tab:phoenix}
\end{table}

\subsection{Implementation details} \label{subsec:exp:implementation}
\noindent \textbf{Features.}
For all datasets, we employ an I3D~\cite{carreira2017quo} backbone architecture, $\phi$, 
pretrained for the task of action recognition on the Kinetics dataset~\cite{kay2017kinetics} to provide representations that are sensitive to fine-grained human motions. To ensure viable adaptation on any unlabelled data, we do not use any annotations from the target domain; therefore we train the model directly on the class labels of \bslcorpus{}. 

\vspace{0.2cm}
\noindent \textbf{MS-TCN pretraining on the source domain.}
For the source domain training phase described in Sec.~\ref{sec:method:src}, we pretrain
an MS-TCN model~\cite{Farha_2019_CVPR} as our sign segmentation network, $f = \psi \circ \phi $, that further processes the outputs of the I3D feature extractor $\phi$ described above with a segmentation network $\psi$ comprising a stack of temporal convolutional layers.  The model is trained under a per-frame binary-class classification objective.
Similar to \cite{Farha_2019_CVPR},
we use an architectural design of 4 stages with 10 layers in each stage, 64 filters, and 
an Adam optimizer.

\vspace{0.2cm}
\noindent \textbf{\methodNameLong{} implementation details and hyperparameters.}
We adopt the \textit{Pelt}~\cite{killick2012optimal} method, an exact changepoint detection algorithm which exhibits linear average runtime behaviour, as a basis for discovering disfluencies in \latestEdit{signing} from the space of I3D features.  We employ the implementation provided by~\cite{TRUONG2020107299} with an L2 cost function on I3D features exacted with a stride of one frame. We use a Pelt cost penalty term of 100 when using changepoints to modulate \pseudolabel{} \textit{insertion} and \textit{refinement}. Both the bandwidth and matching window \latestEdit{hyperparameters} ($\gamma$ and $\delta$) introduced in Sec.~\ref{sec:method:combined} are set to a value of four frames. 
Since boundaries estimated by the changepoint algorithm span only a single frame, we expand the changepoints to span three frames in terms of sign boundary width \latestEdit{(all videos considered in this work are encoded at 25 fps)}.
This ensures that the boundary estimations are similar in width to the \pseudolabels{} produced by the MS-TCN sign segmentation network.

\vspace{0.2cm}
\noindent \textbf{Training and evaluation.}
To evaluate performance, we report the improvement derived from performing adaptation using unlabelled training set videos from the target domain (\textit{inductive} setting). Since our approach does not require the use of labels on the target dataset, we also report results for experiments in which perform adaptation directly on the test data without labels (\textit{transductive setting}), \latestEdit{a formulation which can often arise in practice when the target videos to be segmented are known in advance.}
\latestEdit{Due to the small portion of \bslonek{} with annotations available, we do not use a separate validation set, and fix the number of epochs to 10.}

\begin{table}
    \centering
    \resizebox{0.6\linewidth}{!}{
    \begin{tabular}{lcc}
         \toprule
         PL threshold  &  mF1B & mF1S \\
        \midrule
         0.5 &  $47.94_{\pm1.0}$ &  $32.45_{\pm0.3}$ \\
         0.4  & $48.01_{\pm1.0}$ &  $31.26_{\pm0.7}$ \\
         0.3  & $48.21_{\pm1.0}$ &  $29.87_{\pm1.1}$ \\
         0.2  & $46.67_{\pm0.6}$ &  $27.52_{\pm0.7}$ \\
         \bottomrule
    \end{tabular}
    }
    \vspace{6pt}
    \caption{\textbf{\Pseudolabel{} training with reduced thresholds on the PHOENIX14 dataset:} Reducing the threshold at which a \pseudolabel{} probability is mapped to a boundary label does not address the under-segmentation issue in a way that improves performance. 
    }
    \mbox{}\vspace{-0.4cm} \\
    \label{tab:PL_th}
\end{table}
\begin{table}
    \centering
    \resizebox{0.99\linewidth}{!}{
    \begin{tabular}{lrccc}
         \toprule
          & Adaptation protocol & mF1B & mF1S   \\
         \midrule
         \textit{Baselines:} \\
         \midrule
         Uniform (using GT \#signs)& - & 41.80 & 34.75 \\ 
         Changepoints (using GT \#signs) & - &  60.73 & 52.89\\
         Changepoints (estimating \#signs) & - & 60.25 & 53.94 \\
         \midrule
         \textit{Prior works:} \\
         \midrule
         Geometric features + RF~\cite{farag2019learning, renz20segments} & source-only &  $51.26_{\pm0.5}$ & $34.28_{\pm1.0}$ \\
         MS-TCN~\cite{Farha_2019_CVPR, renz20segments} & source-only  & $61.12_{\pm0.9}$ & $49.96_{\pm0.6}$  \\
         \midrule
         \textit{Proposed model:} \\
         \midrule
         \methodName{} & inductive & $65.99_{\pm1.0}$& $48.81_{\pm1.3}$\\
         \methodName{} &  transductive & $\mathbf{67.01_{\pm2.2}}$& $50.20_{\pm0.6}$ \\
         \bottomrule
    \end{tabular}
    }
    \vspace{6pt}
    \caption{\textbf{Results on the \bslonek{} dataset:} We compare our
    approach to several baselines and the previous state of the art.
    The changepoint-only approach already builds a strong baseline, especially for the mF1S metric. With \methodName{} we gain about 6\% mF1B over the naive transfer from \bslcorpus{}-labelled training  of~\cite{renz20segments}.
    }
    \mbox{}\vspace{-0.7cm} \\
    \label{tab:ablations}
\end{table}

\subsection{Ablation studies} \label{subsec:exp:ablations}

\noindent \textbf{PHOENIX14}. To assess the influence of components of our framework, we first present an ablation study on the PHOENIX14 dataset in Tab.~\ref{tab:phoenix}, in which we compare against the baseline performance of \cite{renz20segments} trained only on source data. By using the \pseudolabelling{} technique, we obtain a first small improvement
(46.75 vs 47.94 mF1B).
It is interesting to note that
the changepoint detection algorithm alone establishes a
very strong baseline (48.51 mF1B).
More generally we observe that self-training with both \pseudolabels{} and changepoints
provide a significant boost in adapting the model to the new domain under both
inductive and transductive model evaluations. However, we see that our \methodNameLong{} (\methodName) provides a significantly greater boost in mF1B (of about 6 points). \\
We also show that the under-segmentation issue is not addressed by simply reducing the threshold employed to convert \pseudolabel{} posterior probabilities into boundary class labels (for all other experiments we simply convert each frame that is assigned a boundary probability higher than 0.5 to a boundary label). Tab.~\ref{tab:PL_th} reports the performance of the model under the influence of varying this threshold.
%
\latestEdit{See
\if\sepappendix1{Sec.~A}
\else{Sec.~\ref{app:sec:ablation}}
\fi
of the appendix
for further ablations on PHOENIX14.
}

\begin{figure}[t]
    \centering
    \includegraphics[width=0.46\textwidth]{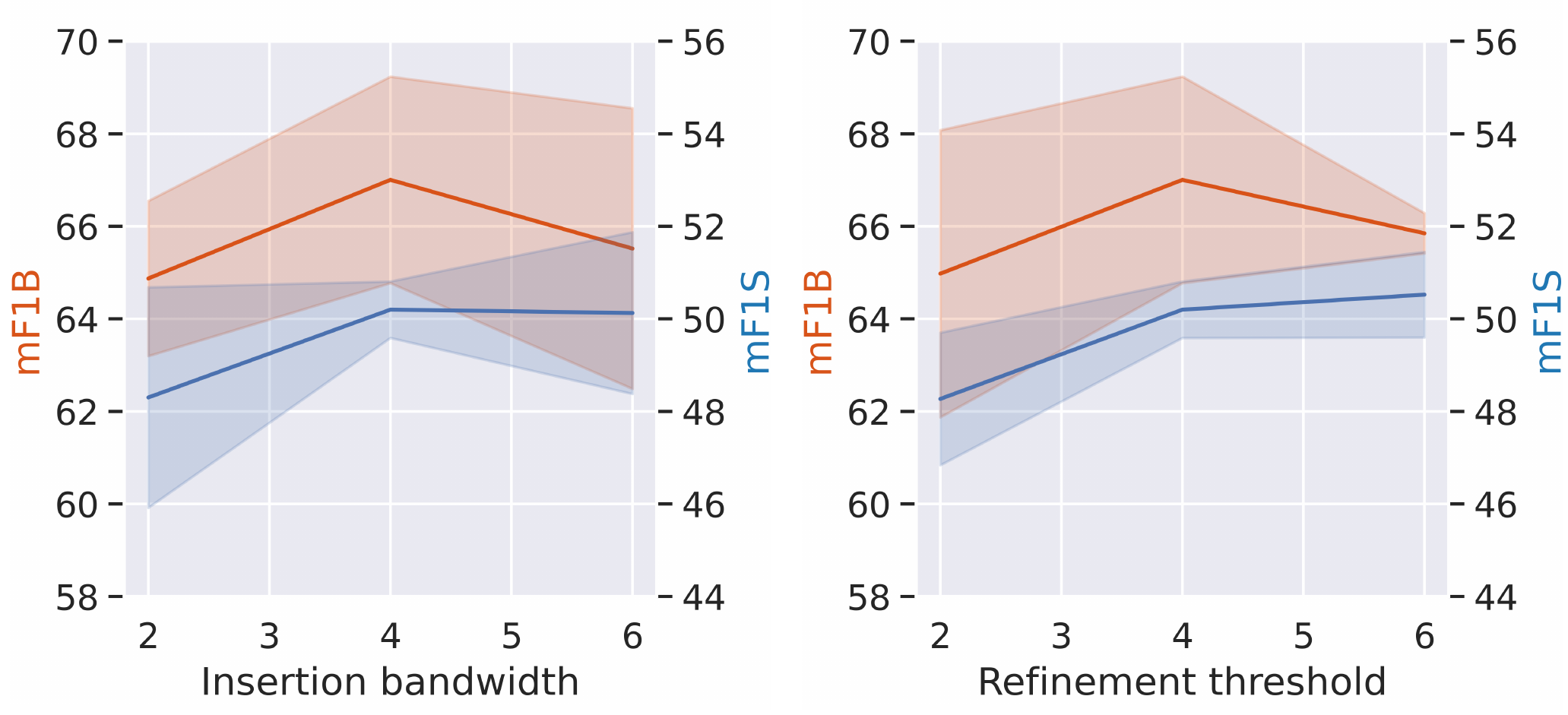}
    \caption{\textbf{Sensitivity to the threshold hyperparameters of the \methodName{} method for \bslonek{} (transductive):} We show the influence of different values for thresholds for the \textit{insertion} bandwidth $\gamma$ (left) and \textit{refinement} matching window $\delta$ (right). 
    }
    \mbox{}\vspace{-0.7cm} \\
    \label{fig:cmsl_param_th}
\end{figure}

\begin{figure}
    \centering
    \includegraphics[width=0.44\textwidth]{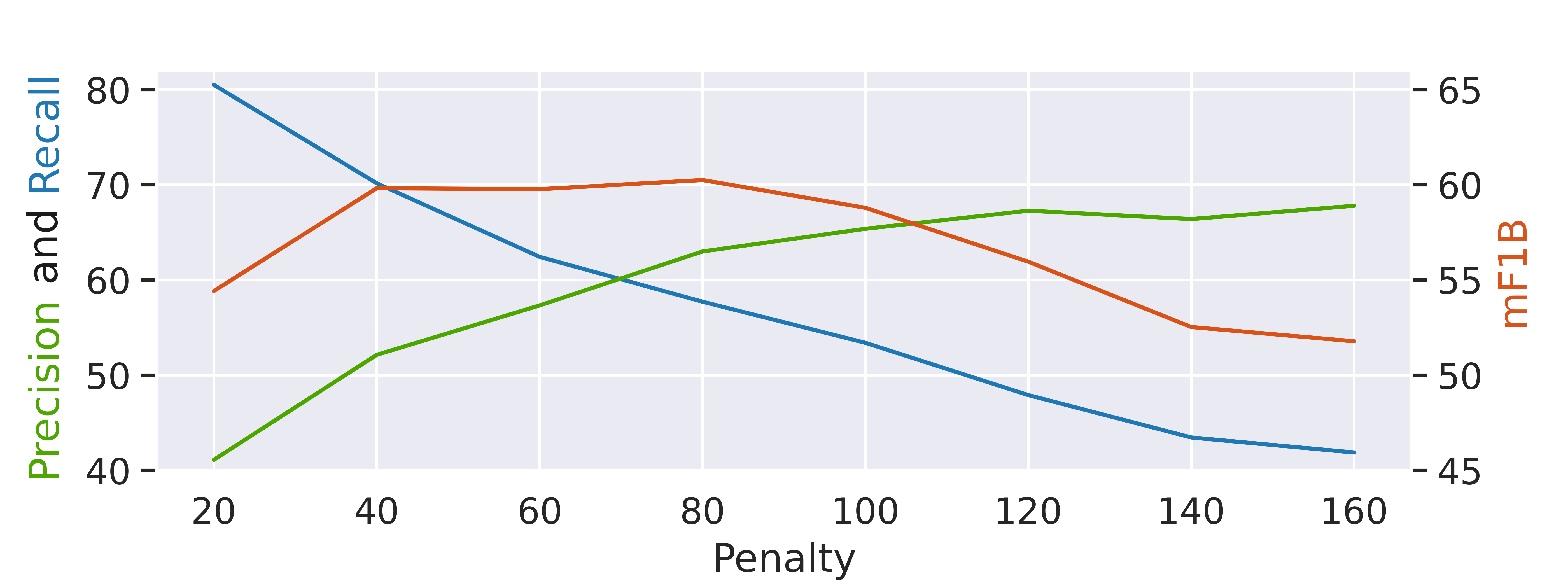}
    \caption{\textbf{Pelt changepoint penalty term:} We study the influence of the penalty value for Pelt changepoint detection algorithm when using only changepoints as labels to train 
    on the target domain. We observe that high penalty results in under
    segmentation, whereas a low penalty score over-segments. 
    }
    \mbox{}\vspace{-0.9cm} \\
    \label{fig:penalty}
\end{figure}

\vspace{0.2cm}
\noindent \textbf{BSL-1K.} 
We observe that in both the inductive and transductive setting, \methodName{} yields a significant gain over the strongest reported result for sign segmentation on \bslonek. In particular, we observe an improvement of about 6\% mF1B over the naive transfer of the model trained on the source data.
To better understand the behaviour of the proposed approach, we next conduct further ablations using various baselines on the \bslonek{} dataset.  We describe these baselines next and report their performance in Tab.~\ref{tab:ablations}. We also show the performance of alternative fusion strategies and the hyperparamter sensitivity of our approach. \\
\noindent \textit{Uniform baseline.}
We report the performance of uniformly splitting the target video into segments of equal size such that the total number of segments match the total number of ground-truth boundaries. Since the uniform baseline assumes that the ground-truth number of signs are known, it is thus \textit{not} directly comparable to the automatic segmentation models, but provides a degree of calibration for the difficulty of the task. \\
\noindent \textit{Changepoints-only baseline.}
For the changepoint baseline, also reported in Tab.~\ref{tab:ablations}, we employ only the feature extraction network, $\phi$, and calculate the changepoints on the extracted features, $\phi(\mathbf{x})$. The first method uses dynamic programming with the ground-truth number of boundaries given.
By using the changepoint algorithm, we observe a large improvement in both metrics over uniformly splitting the video.
However, we also observe that by using the Pelt method to predict position and number of boundaries we come close to matching the variant that uses ground truth information, suggesting the robustness of this technique and its suitability as a building block in our approach. In Fig.~\ref{fig:penalty}, we assess the sensitivity of this approach to the choice of different values for the penalty hyperparameter and determine that a choice of 80 represents a reasonable trade-off for the changepoints-only baseline.\\
\begin{figure}[t]
    \centering
    \includegraphics[width=0.46\textwidth]{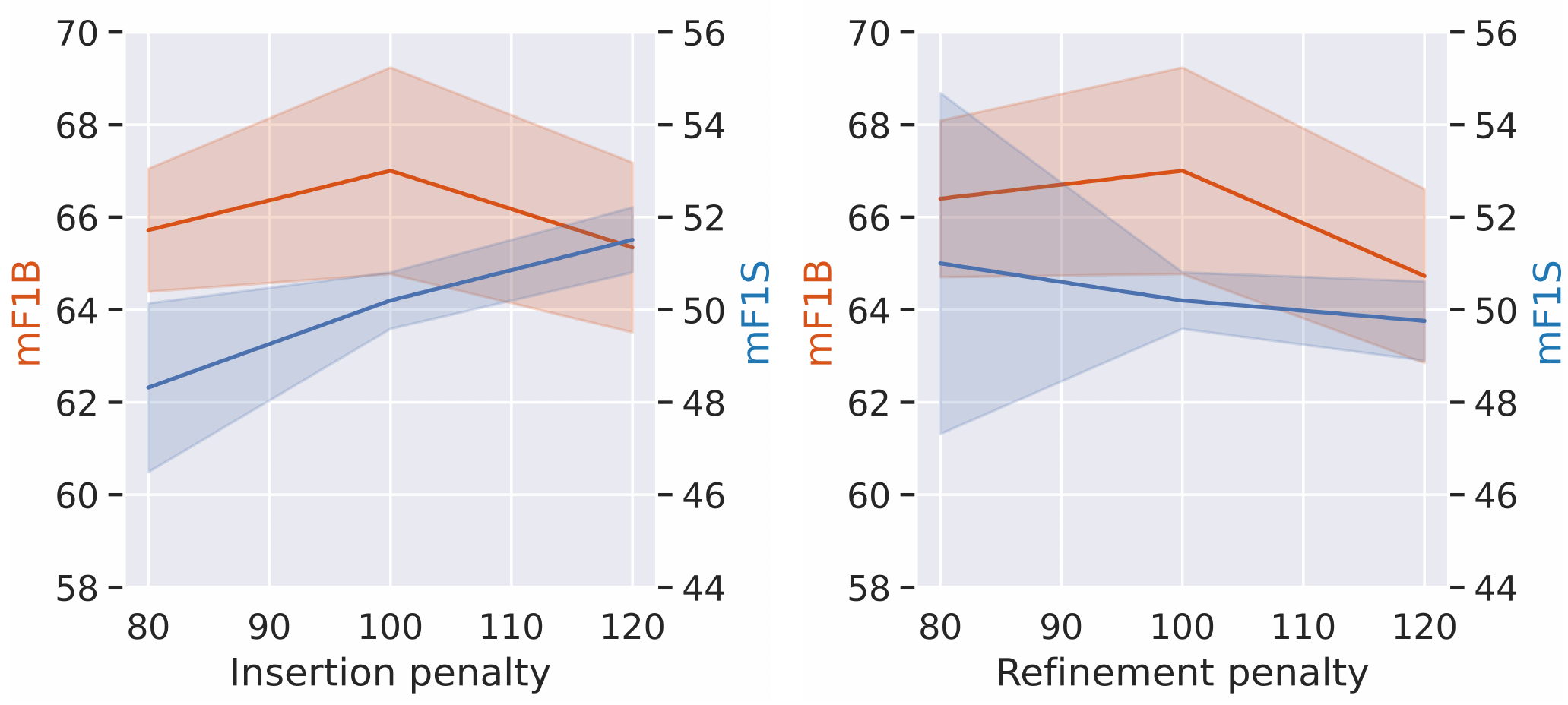}
    \caption{\textbf{Sensitivity to the penalty hyperparameter of the \methodName{} method for \bslonek{} (transductive):} We show the influence of different values for varying the changepoint penalty when fixing the threshold hyperparameters to $\gamma=4$, $\delta=4$ as used in all other experiments. 
    }
    \mbox{}\vspace{-0.7cm} \\
    \label{fig:cmsl_param_pen}
\end{figure}
\begin{table}
    \centering
    \resizebox{0.99\linewidth}{!}{
    \begin{tabular}{lcrr}
         \toprule
         Fusion strategies & Adaptation protocol & mF1B & mF1S \\
         \midrule
         merge PL + CP  & inductive &  $65.10_{\pm1.6}$& $42.73_{\pm1.0}$  \\
         local fusion & inductive &  $62.35_{\pm1.4}$& $48.77_{\pm2.3}$  \\
         \midrule
         insertion & inductive & $62.49_{\pm2.2}$& $45.77_{\pm1.6}$\\
         insertion + refinement  (\methodName{}) & inductive & $\mathbf{65.99_{\pm1.0}}$& $48.81_{\pm1.3}$\\ 
         \midrule
         \midrule
         merge PL + CP & transductive & $65.69_{\pm1.3}$& $43.06_{\pm3.0}$ \\
         local fusion & transductive & $62.71_{\pm3.1}$& $50.69_{\pm1.9}$ \\
         \midrule
         insertion & transductive & $63.27_{\pm3.4}$& $48.49_{\pm3.1}$ \\
         insertion + refinement (\methodName{}) & transductive & $\mathbf{67.01_{\pm2.2}}$& $50.20_{\pm0.6}$ \\
         \bottomrule
    \end{tabular}
    }
    \vspace{6pt}
    \caption{\textbf{Fusion strategies:} We compare several strategies for fusing \pseudolabels{} and changepoints on the \bslonek{} test set. We observe that the combined ``insertion + refinement'' strategy proposed as part of the \methodName{} approach consistently performs best.}
    \mbox{}\vspace{-0.7cm} \\
    \label{tab:fusion}
\end{table}
\noindent\hspace{-7pt}\textit{Effect of the fusion strategy.} We explore several alternative ways to fuse the two knowledge sources provided by \pseudolabels{} and changepoint detections. The simplest such strategy takes the union of predictions and keeps all boundaries which are present either in the \pseudolabels{} or in the changepoints (labelled \say{merge PL + CP} in Tab.~\ref{tab:fusion}).  We also consider an alternative strategy that specifically targets the under-segmentation problem induced by a \pseudolabel-only approach. To insert more boundaries we select those segments which are longer than the average sign length and insert detected changepoints only amongst such regions. These results are reported in Tab.~\ref{tab:fusion} under the row title \say{local fusion}. Finally, we evaluate our proposed \methodName{} fusion strategy and investigate the impact of the two different \pseudolabel{} transformations that it employs.
In Tab.~\ref{tab:fusion}, we observe when used in isolation the \textit{insertion} transformations produce some improvement, but that there is a clear benefit to the combination of \textit{insertion} and \textit{refinement}, justifying the usage of this combination in our design. \\
\noindent\textit{Hyperparameter sensitivity.}
We further show the sensitivity to the hyperparameters of the \methodName{} method. First we show the impact of different values for thresholds for the \textit{insertion} bandwidth $\gamma$ and \textit{refinement} matching window $\delta$ in Fig.~\ref{fig:cmsl_param_th}.
Changes to these hyperparameters produce a small variation in the results. We observe that an insertion bandwidth of $\gamma=4$ frames and a refinement matching window of $\delta=4$ frames with the given changepoint penalty of 100 works well.
Next, in Fig.~\ref{fig:cmsl_param_pen}, we fix  $\gamma=4$ and  $\delta=4$ and change the changepoint penalty. We observe the best result in the mF1B metric for a penalty of 100. Looking at the mF1S metric, leads to the assumption, that this metric gains from the insertion of fewer boundaries.


\begin{figure}[t]
    \centering
    \includegraphics[width=0.48\textwidth]{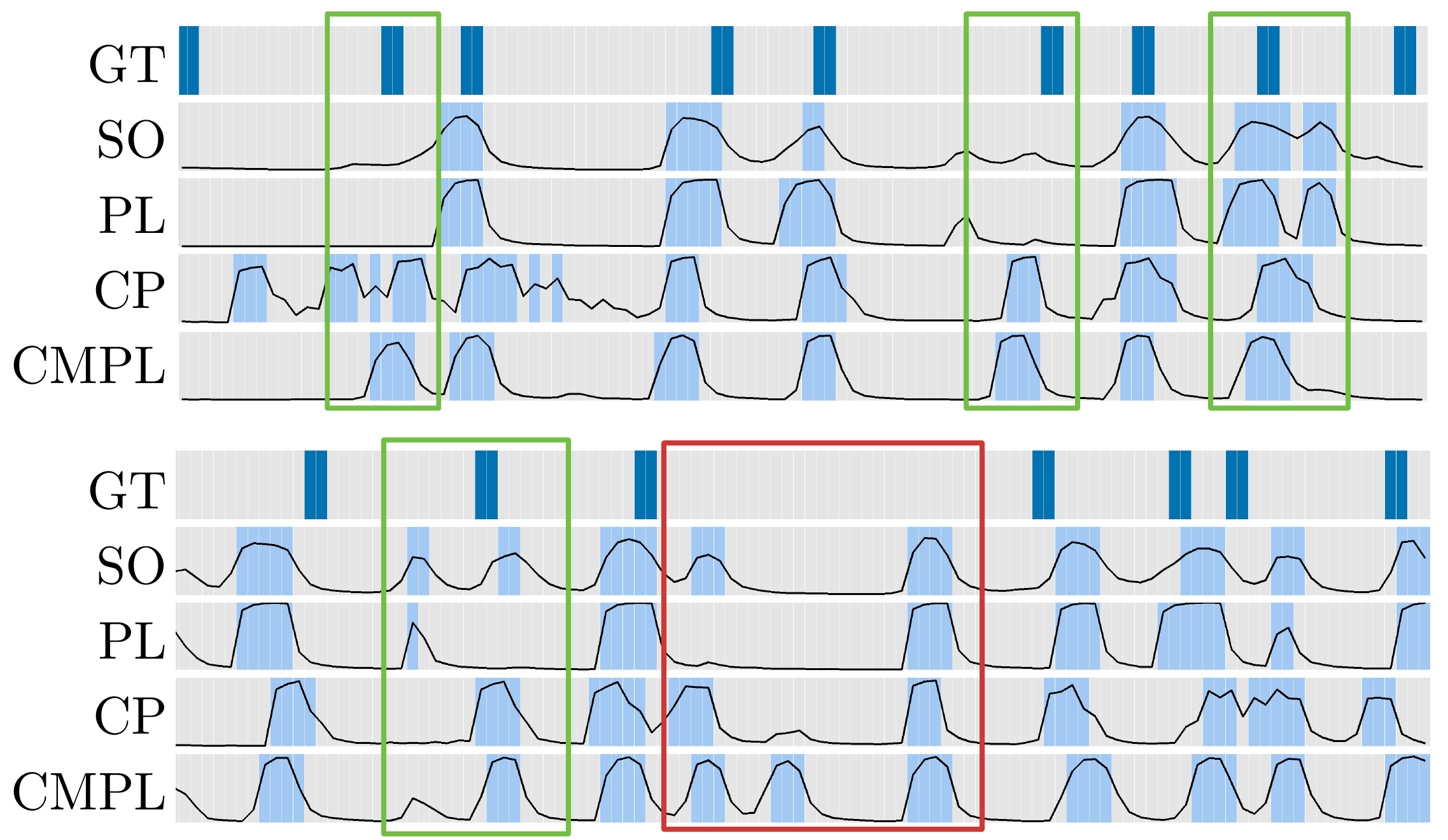}
    \caption{
    \textbf{Qualitative results on PHOENIX14~\cite{Koller15cslr}:}
    We compare the results of the different models, source-only (SO), \pseudolabels{} (PL), changepoints (CP), \methodNameLong{}{} (\methodName{}) with the ground truth (GT).
    We show two different extracts, in which our method is able to detect new and suppress wrong boundaries (green box). The red box indicates failure cases, where further work is necessary.
    }
    \mbox{}\vspace{-0.9cm} \\
    \label{fig:qualitative:phoenix}
\end{figure}


\subsection{Qualitative analysis} \label{subsec:qual}
\begin{figure}[t]
    \centering
    \includegraphics[width=0.438\textwidth]{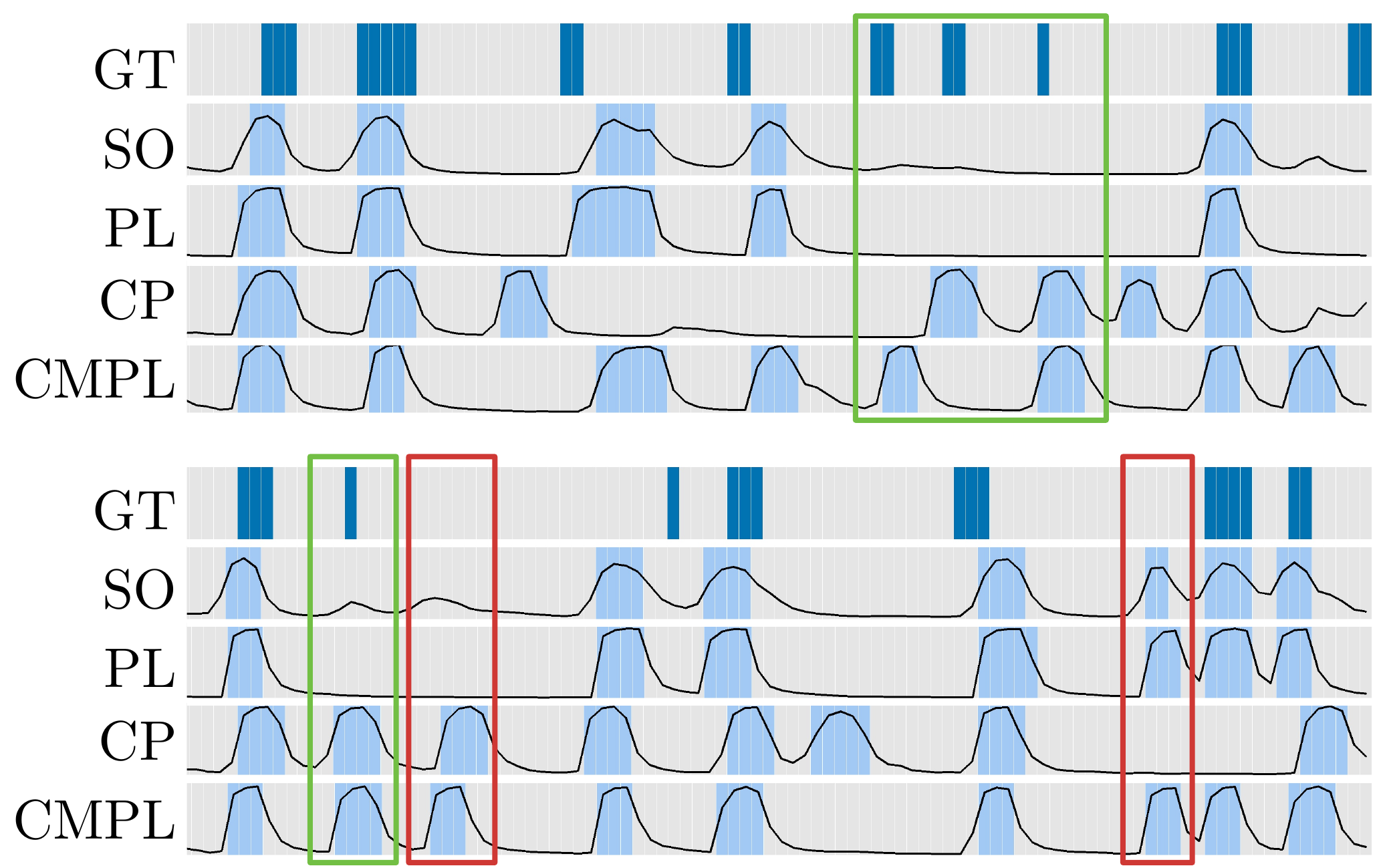}
    \caption{
    \textbf{Qualitative results on \bslonek{}~\cite{albanie20_bsl1k}:}
    (Top) Changepoints help to find boundaries which were previously missed. (Bottom) Through the insertion of the changepoint the \methodName{} method detects new boundaries, but is in most cases not able to suppress the false positives detected by the source-only approach.
    }
    \mbox{}\vspace{-0.75cm} \\
    \label{fig:qualitative:bsl1k}
\end{figure}

In Fig.~\ref{fig:fusion}, we illustrate two examples of the proposed \methodNameLong{} method in action. The upper one depicts an extract of an episode in which the described under-segmentation problem of the \pseudolabels{} (PL) can be clearly observed. As the changepoint detection finds a boundary and no \pseudolabel{} is in the neighbouring area, we insert a new boundary at the detected position, remedying the omission.
In the lower example, we observe the behaviour of the refinement transformation and its role in improving overall \pseudolabel{} quality by adjusting the boundary locations to account for evidence from the changepoint detections. \\
In Fig.~\ref{fig:qualitative:phoenix} and Fig.~\ref{fig:qualitative:bsl1k} we provide a qualitative comparison of the different models on two sequences from the PHOENIX14 and \bslonek{} datasets. We show failure cases in red and success cases in green.
In Fig.~\ref{fig:qualitative:phoenix}, the upper example demonstrates the effectiveness of the \methodName{} for predicting new boundaries through the changepoint insertion. The first two false positive boundaries in the bottom example are due to a repeated sign, which leads to abrupt changes at the time the sign gets repeated. 
The last boundary in the red frame, which is falsely detected by each method, is aligned with a strong direction change in the signing.
In general, we observed in our error analysis two main issues: (i) the new method mainly helps against under-segmentation, but can only in seldom cases suppress false positive predictions in the \pseudolabels{} or changepoints, (ii) the insertion of the changepoints leads to higher true positive rate, but also to a higher false positive rate. This can be observed especially for PHOENIX14.
The previously mentioned two issues can be also observed for \bslonek{} (Fig.~\ref{fig:qualitative:bsl1k}). We detect more correct boundaries (green box), but also some false positives (left red box) and are not able to suppress already existing false positives (right red box).
\latestEdit{For further qualitative
results, we refer to our supplemental video
on the project webpage and
\if\sepappendix1{Sec.~B}
\else{Sec.~\ref{app:sec:qualitative}}
\fi
of the appendix.}

%% file: conclusions.tex
\section{Conclusion}
\label{sec:conclusion}

We presented an approach to temporally
segment signs in continuous sign language
videos, with a particular emphasis on
leveraging unlabelled data for training.
We motivated and introduced the problem of \textit{source-free}
domain adaptation in this context, in which we assume access to labelled 
source data during model training but not during adaptation to the target domain.
We proposed the conceptually simple but powerful \methodNameLong{} algorithm
and demonstrated its effectiveness
through state-of-the-art performance
on two sign language datasets.
Potential future directions include
repurposing our segmentation models
for improving sign language recognition
and active signer detection
performance.

%% file: acknowledgements.tex
{\small \noindent\textbf{Acknowledgements.}
This work was supported by EPSRC grant ExTol.
KR was supported by the German Academic Scholarship Foundation.
\latestEdit{The authors thank Andrew Zisserman for suggestions and Cihan Camg\"oz for assistance with data preparation.} }

%% file: appendix.tex
\renewcommand{\thefigure}{A.\arabic{figure}}
\setcounter{figure}{0} 
\renewcommand{\thetable}{A.\arabic{table}}
\setcounter{table}{0} 

\appendix

This document provides further ablations on 
PHOENIX14 (Sec.~\ref{app:sec:ablation}) and \latestEdit{
further interpretation on the results including the supplemental video and a discussion on motion blur} 
(Sec.~\ref{app:sec:qualitative}).

\section{\latestEdit{Additional ablations}} \label{app:sec:ablation}

\noindent\textbf{Effect of the \methodName{} components.}
Tab.~\ref{tab:phoenix_components} shows the influence of the two components of the proposed \methodName{} method. We observe a gain of approximately 5 percent with the insertion of the changepoints. The refinement stage provides a further 1-2 percent of improvement.

\noindent\textbf{Multiple iterations.}
We report results for multiple iterations of \pseudolabel{} training, i.e.,
finetuning with the \pseudolabels{} obtained through the previous iteration.
Tab.~\ref{tab:iteration} summarises the results.
Performance does not improve with more iterations
when using \pseudolabels{} or \methodName{}.

\section{Qualitative results} \label{app:sec:qualitative}

\noindent\textbf{Supplemental video.}
\latestEdit{We refer to the supplemental video
on our project webpage to assess the sign segmentation
performance qualitatively. We identify success and failure
cases on all datasets used in this work.}

\noindent\textbf{Motion blur.}
We conduct a study on the correlation of model performance and motion blur.
For this, we use the number of detected hand keypoints as a proxy for motion blur. The assumption is, that the keypoint detection works better for sharper images. This could be confirmed in a quantitative evaluation. Since our metrics work on a video and not frame-level we average the blur score over the video.
We calculate the correlation coefficient $r$ and only find a very weak correlation with the mF1B score on \bslcorpus{} ($r=0.09$).

\begin{table}[hb]
    \centering
    \resizebox{0.85\linewidth}{!}{
    \begin{tabular}{lrrr}
         \toprule
         Fusion strategies& mF1B & mF1S \\
         \midrule
         Source-only & $46.75_{\pm1.2}$ & $32.29_{\pm0.3}$\\
         \midrule
         adaptation protocol: \textit{inductive}\\
         \midrule
         insertion & $52.01_{\pm0.5}$ &  $32.83_{\pm0.6}$\\ 
         insertion + refinement (\methodName{}) & $\mathbf{53.57_{\pm0.7}}$ &  $\mathbf{33.82_{\pm0.0}}$\\
         \midrule
         adaptation protocol: \textit{transductive}\\
         \midrule
         insertion & $51.37_{\pm0.5}$ &  $32.70_{\pm1.4}$\\ 
         insertion + refinement (\methodName{})   & $\mathbf{53.53_{\pm0.1}}$ &  $\mathbf{32.93_{\pm0.9}}$\\
         \bottomrule
    \end{tabular}
    }
    \vspace{3pt}
    \caption{\textbf{Impact of the two components of 
    \methodName{} 
    on PHOENIX14:} We 
    provide an ablation which shows the impact of the insertion and refinement stage.
    The insertion of the changepoints gives the largest 
    boost. The refinement stage provides further 
    improvement.
    }
    \label{tab:phoenix_components}
\end{table}

\begin{table}[hb]
    \centering
    \resizebox{0.65\linewidth}{!}{
    \begin{tabular}{lrr}
         \toprule
         Iteration & mF1B & mF1S \\
         \midrule
         PL 1 & $47.94_{\pm1.0}$ &  $32.45_{\pm0.3}$ \\
         PL 2 & $47.79_{\pm0.7}$ &  $32.10_{\pm0.4}$ \\
         PL 3 & $47.93_{\pm0.9}$ &  $31.80_{\pm0.2}$ \\
         \midrule
         \methodName{} 1 & $\mathbf{53.57_{\pm0.7}}$ &  $33.82_{\pm0.0}$\\
         \methodName{} 2 & $53.56_{\pm1.1}$ &  $34.27_{\pm0.7}$\\
         \bottomrule
    \end{tabular}
    }
    \vspace{3pt}
    \caption{\textbf{Multiple iterations of pseudo training on PHOENIX14:} We show that multiple iterations improve the results for training with only \pseudolabels{} but can not reach the performance of \methodName{} for which only one iteration is sufficient. 
    }
    \label{tab:iteration}
\end{table}





